\documentclass[11pt]{article}
\usepackage[final]{acl}
\usepackage{times}
\usepackage{latexsym}
\usepackage{enumitem}
\usepackage{pifont}
\usepackage[T1]{fontenc}
\usepackage{placeins}
\usepackage{enumitem}
\usepackage{multirow}
\usepackage{tabularx}
\usepackage[linesnumbered,ruled,vlined]{algorithm2e}
\usepackage{booktabs}      
\usepackage{threeparttable} 
\usepackage{amssymb}       
\usepackage{graphicx}   
\usepackage{subcaption} 
\usepackage{tikz}       
\usepackage{pgfplots}   
\pgfplotsset{compat=1.18}
\usepgfplotslibrary{groupplots} 
\usepackage[english,bidi=default]{babel} 
\usetikzlibrary{plotmarks}
\usetikzlibrary{patterns}
\usepackage{amsmath}
\usepackage{soul} 
\usepackage[linesnumbered,ruled,vlined]{algorithm2e}
\usepackage{booktabs}      

\usepackage{microtype}

\usepackage[utf8]{inputenc}

\newcommand{\softbsubsec}[1]{\vspace{0.5em}\noindent\textbf{#1.}}

\definecolor{deltlight}{RGB}{144,190,109}
\definecolor{deltmed}{RGB}{76,153,0}
\definecolor{deltdark}{RGB}{0,100,0}
\newcommand{\deltL}[1]{{\scriptsize\color{deltlight}#1}}
\newcommand{\deltM}[1]{{\scriptsize\color{deltmed}#1}}
\newcommand{\deltH}[1]{{\scriptsize\color{deltdark}#1}}

\usepackage{microtype}

\usepackage{inconsolata}

\usepackage{graphicx}

\usepackage{xspace}

\definecolor{ACLLinkBlue}{HTML}{000099}
\hypersetup{colorlinks=true, allcolors=ACLLinkBlue}

\newcommand{\sys}{{XoG}\xspace}

\usepackage{fancyhdr}

\fancypagestyle{plain}{
    \fancyhf{}
    \cfoot{\thepage}

}

\begin{document}
\thispagestyle{plain}
\clearpage
\newgeometry{top=0.75in,bottom=0.75in,left=0.85in,right=0.85in}

\nolinenumbers
\small
\normalsize

\clearpage
\restoregeometry

\title{Explore-over-Graph: Hybrid Embedding–LLM Reasoning for Knowledge Graph Question Answering under Incompleteness}

\author{Ola El Khatib \quad Djellel Difallah \\
  New York University Abu Dhabi, UAE \\
  \texttt{\{oge208,djellel\}@nyu.edu}}

\maketitle

\begin{abstract}Large language models (LLMs) are increasingly combined with knowledge graphs (KGs) to ground reasoning in structured evidence. However, most LLM-based KGQA methods rely on traversing existing graph edges and become unreliable when reasoning paths are broken by missing facts. Alternatives that ask LLMs to generate missing knowledge risk introducing hallucinated evidence. We introduce \sys (eXplore-over-Graph), a framework for multi-hop question answering over incomplete KGs that recovers missing reasoning paths from learned graph structure rather than LLM parametric knowledge. \sys combines type-level entity–relation statistics to identify candidate relations with KG embeddings to retrieve plausible missing entities, using the LLM as a semantic selector and reasoner. These mechanisms are integrated into an iterative planning–exploration–reasoning process. Experiments on WebQSP, CWQ, and the Wikidata-based BRINK benchmark show that \sys remains competitive on complete KGs and consistently outperforms comparable methods without task-specific KGQA training under KG incompleteness. These gains persist across multiple LLM backbones, indicating that stronger LLMs alone do not resolve missing graph evidence. \sys also reduces LLM token consumption by up to 33\% compared with a closely related planning-based approach.

\end{abstract}

\begin{figure}[htb!]
    \centering
    \includegraphics[width=\columnwidth]{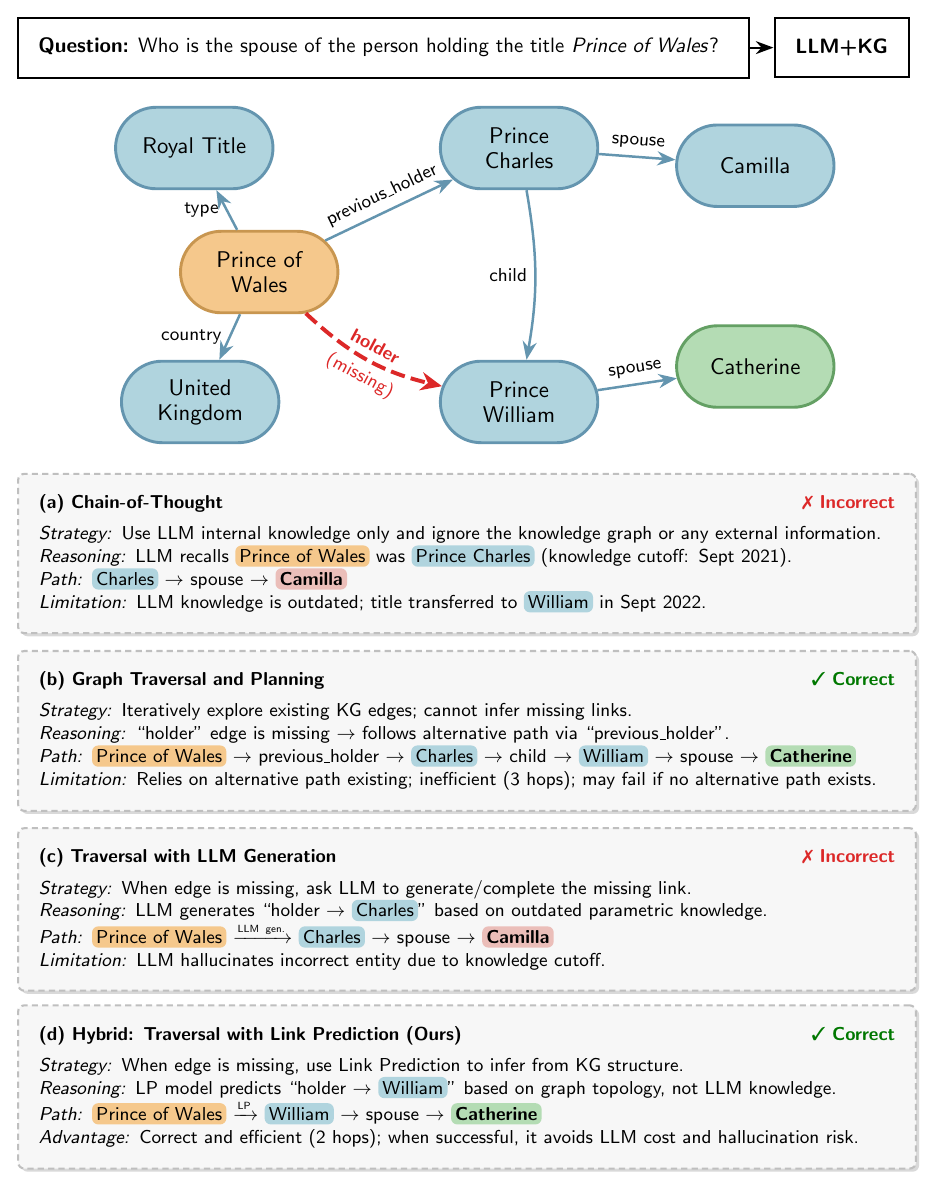}
    \caption{Comparison of LLM+KG reasoning paradigms.}
    \label{fig:motivation}
\end{figure}

\section{Introduction}\label{sec:intro}

LLMs have demonstrated remarkable success across a wide range of natural language processing tasks, including question answering and complex reasoning~\cite{wei2022chain,yao2023react}, but remain prone to hallucinations and out-of-date knowledge due to static parametric memory, which poses fundamental limitations for knowledge-intensive and multi-hop reasoning tasks ~\cite{DBLP:journals/tois/HuangYMZFWCPFQL25}. To mitigate these limitations, recent research integrates LLMs with KGs~\cite{ji2021survey}, combining the structured and verifiable knowledge provided by KGs with the language understanding and reasoning capabilities of LLMs. However, real-world KGs are inherently incomplete due to their scale, dynamic nature, and long-tail relation distributions, with missing entities and relations that hinder multi-hop reasoning. We refer to this setting as KG incompleteness, where one or more triples required to form a correct multi-hop reasoning path are missing from the graph. Such incompleteness poses a fundamental challenge for knowledge graph question answering (KGQA).

Existing work on LLM-based KGQA uses LLMs to guide interactive graph exploration~\cite{sun2023think}, but their performance can degrade substantially when required facts are missing. Generation-based methods address this limitation by generating missing triples during exploration, but may introduce incorrect information~\cite{xu2024generate}. Fig.~\ref{fig:motivation} illustrates these limitations for question answering over an incomplete KG: pure chain-of-thought (CoT) reasoning relies on the LLM's internal knowledge (Fig.~\ref{fig:motivation}(a)); planning-based approaches may compensate for broken paths by traversing longer routes at the cost of increased exploration depth (Fig.~\ref{fig:motivation}(b)); and generation-based methods generate missing triples directly but may introduce outdated or incorrect facts (Fig.~\ref{fig:motivation}(c)). These limitations reveal a trade-off between \emph{efficiency} and \emph{reliability} under KG incompleteness (Fig.~\ref{fig:motivation}(d)).

Motivated by these challenges, we introduce \sys, a multi-hop KGQA framework that addresses KG incompleteness through KG-grounded recovery. \sys reconnects broken reasoning paths while constraining the LLM to select from curated candidate sets. Specifically, \sys uses (a) co-occurrence statistics between entity types and relations to prune the large relation search space, and (b) embedding-based link prediction to retrieve likely target entities, with language models acting as a semantic selector over noisy candidates. These are integrated into an iterative planning--exploration--reasoning strategy that combines query decomposition, graph exploration, and final answer reasoning. Experiments on benchmark KGQA datasets show that \sys improves answer accuracy and token efficiency under KG incompleteness while remaining competitive in complete-KG settings.

In summary, our main contributions are:
\begin{itemize}[itemsep=0pt,topsep=0pt,leftmargin=*,label=$\star$]

\item We introduce \sys, a multi-hop KGQA framework that addresses KG incompleteness through two complementary KG-grounded recovery mechanisms: type-level entity--relation co-occurrence statistics to identify likely relations given context, and embedding-based link prediction to retrieve the tail entities that ground LLM reasoning.\looseness-2

\item To control the candidate-expansion noise introduced by recovery, we propose a hierarchical bucketing strategy that scales LLM relation pruning to KGs with thousands of relations and use DistilBERT-based semantic filtering for entity candidates, all within an iterative \emph{planning--exploration--reasoning} loop rather than as one-shot graph completion.\looseness-2

\item We conduct extensive experiments on WebQSP and CWQ under varying degrees of KG incompleteness, and validate generalization on the Wikidata-based BRINK benchmark, demonstrating that \sys is competitive with state-of-the-art methods on complete graphs and consistently strongest among methods without task-specific KGQA training under incompleteness, while improving token efficiency.\looseness-2

\end{itemize}
\section{Related Work}\label{sec:related-work}

\softbsubsec{KGQA under incomplete KGs}
Early KGQA approaches relied on semantic parsing to convert questions into executable logical forms, requiring complete and well-curated knowledge graphs.
To mitigate KG incompleteness, embedding-based methods recover missing entities or relations via similarity scores in learned vector spaces (e.g., EmbedKGQA~\cite{saxena2020improving}, Query2Box~\cite{ren2020query2box}, LEGO~\cite{pmlr-v139-ren21a}, BeamQA~\cite{atif2023beamqa}).
While effective, these approaches depend heavily on training data quality and often introduce noise in long-tail settings.
Other works address incompleteness through relation prediction~\cite{zhao2022improving} or multi-level knowledge generation~\cite{ijcai2024p236}, while Var2Vec~\cite{DBLP:conf/aaai/WangCG23} embeds logical variables alongside link prediction for efficient query answering.

\softbsubsec{KGQA with LLM Reasoning}
Advances in LLM reasoning through prompting strategies, including Chain-of-Thought~\cite{wei2022chain} and its variants~\cite{zhang2022automatic,fu2022complexity,wang2022self,kojima2022large,sun2024enhancing,yao2024tree,besta2024graph}, decomposition-based methods~\cite{khot2022decomposed}, and ReAct~\cite{yao2023react}, have enabled integration of LLMs into KGQA pipelines. ToG~\cite{sun2023think} and PoG~\cite{chen2024plan} guide LLMs to iteratively explore KG evidence, but assume complete graphs. GoG~\cite{xu2024generate} relaxes this by allowing LLMs to infer missing links from parametric knowledge, at the cost of increased hallucination risk, while RoG~\cite{luo2023reasoning} relies on fine-tuning. Other graph-enhanced LLM approaches, including G-Retriever~\cite{he2024gretriever} and GNN-RAG~\cite{mavromatis2024gnnrag}, integrate trainable graph retrieval or graph neural modules with LLM reasoning, requiring additional training or adaptation components. Overall, existing approaches either depend on complete KGs, require additional task-specific training, or remain vulnerable to hallucination under KG incompleteness.

\section{Methodology}

\subsection{Problem Statement}

A \emph{knowledge graph} $\mathcal{G} = (\mathcal{E}, \mathcal{R}, \mathcal{T})$ is a directed labeled graph, where each triple $(h, r, t) \in \mathcal{T} \subseteq \mathcal{E} \times \mathcal{R} \times \mathcal{E}$ represents a relation $r$ from head entity $h$ to tail entity $t$.

Knowledge graph embedding (KGE) models represent entities and relations in a low dimensional vector space and measure the likelihood of a triple $(h,r,t)$ via a scoring function $\phi: \mathcal{E} \times \mathcal{R} \times \mathcal{E} \rightarrow \mathbb{R}$. In this work, the scoring function is instantiated using the ComplEx model~\cite{TrouillonWRGB16}, which embeds entities and relations in a complex vector space $\mathbb{C}^d$ and is capable of modeling asymmetric relations.

Knowledge graph question answering (KGQA) aims to answer a query $Q$ given a topic entity $T$ by retrieving answer entities $W \subset \mathcal{E}$. Answers are typically reached through multi-hop reasoning chains such as: $e_0 \xrightarrow{r_1} e_1 \xrightarrow{r_2} \cdots \xrightarrow{r_n} e_n$.

We consider \emph{KGQA under incompleteness}, where the knowledge graph $\mathcal{G}$ may miss facts required for correct reasoning. In particular, some triples along such a chain may be missing, i.e., for some indices $i$, $(e_{i-1}, r_i, e_i) \notin \mathcal{T}$. Our goal is therefore to recover missing triples using the observed graph structure and learned embeddings.

\subsection{Framework Overview}
\begin{figure}[ht]
  \centering
  \includegraphics[width=\columnwidth]{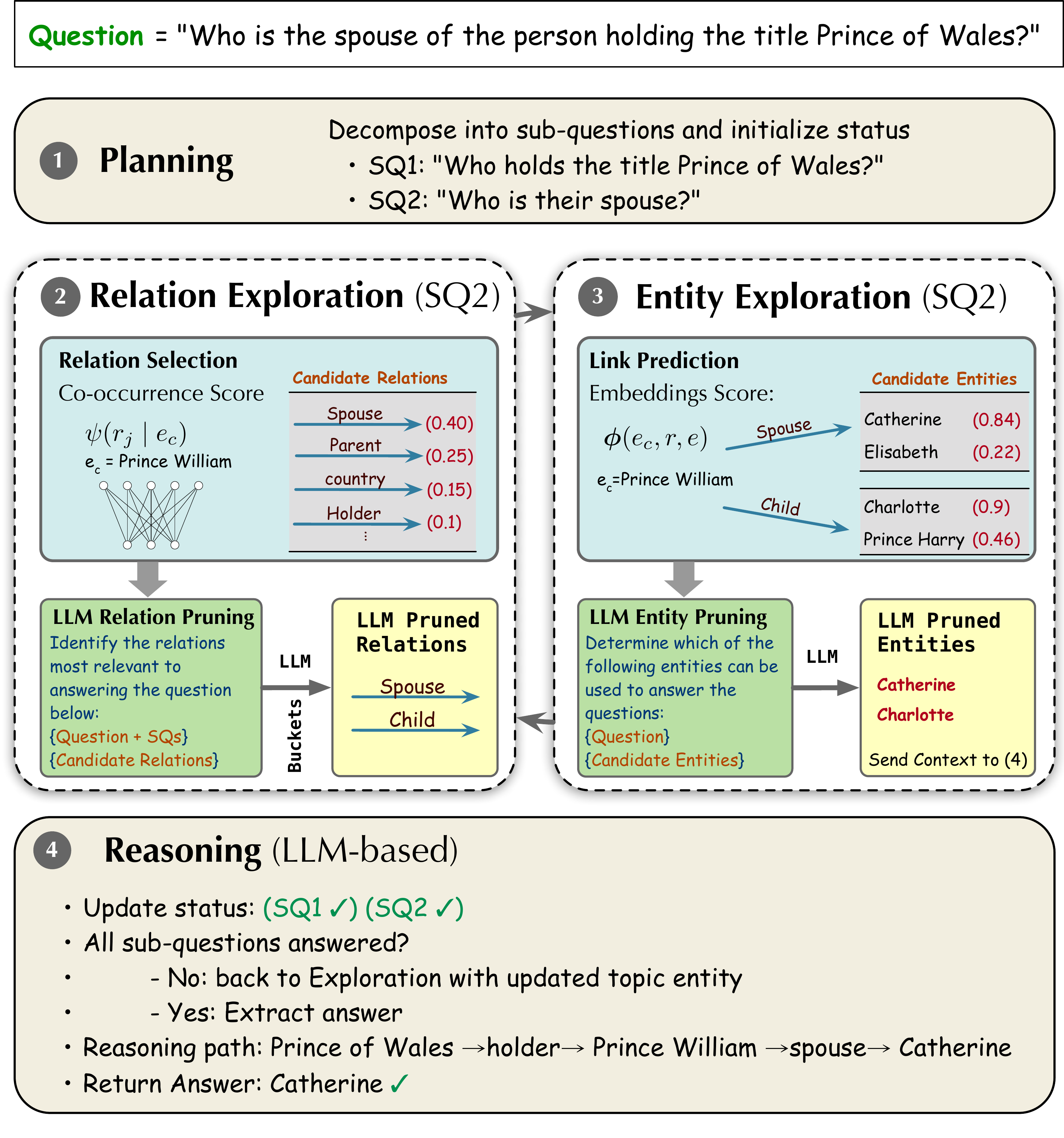}
  \caption{The iterative exploration-reasoning process in \sys for answering a KG-based question.}
  \label{fig:sys_steps_2}
\end{figure}

We propose a modular framework for KGQA designed to be robust under KG incompleteness. As summarized in Fig.~\ref{fig:sys_steps_2}, \sys follows an iterative \emph{planning-exploration-reasoning} paradigm. In the \textbf{planning stage}, the input question is decomposed into sub-questions whose completion status is explicitly tracked throughout the process. The \textbf{ exploration stage} consists of two components: (i) \emph{relation exploration}, which proposes relations relevant to the current reasoning state, and (ii) \emph{entity exploration}, which retrieves plausible next-hop entities, including those inferred via embeddings. In the \textbf{reasoning stage}, an LLM evaluates whether the explored paths are sufficient to answer the question. If not, \sys iterates back to the exploration stage to further expand the paths; otherwise, the answer is extracted directly from the explored evidence.

To handle incompleteness, \sys incorporates recovery mechanisms: relation exploration leverages a curated candidate set derived from corpus-level statistics, while entity exploration uses KG embeddings and link prediction to infer missing entities. Throughout this process, the LLM acts as a guiding agent, pruning noisy candidates and steering traversal based on the reasoning context.\looseness-2

\subsection{Planning}

To guide adaptive exploration and reasoning, \sys decomposes the input question into a set of sub-questions that capture its underlying question semantics. This decomposition is produced by LLM prompting to generate intermediate questions. Given a question $Q$, the resulting sub-questions $\{SQ_i\}$ form an explicit execution plan that directs subsequent exploration and reasoning steps. 

Following \citet{chen2024plan}, \sys maintains a \emph{sub-question status} for each $SQ_i$, indicating whether it has been resolved based on the reasoning paths explored so far. All sub-questions are initially marked as unanswered, and their status is updated iteratively as new paths are explored.

\subsection{Relation and Entity Exploration}

The exploration phase expands the search over the KG by retrieving relevant facts across multiple hops from the topic entity. Unlike purely traversal-based approaches, \sys augments exploration by inferring missing relations and entities from graph structure and learned embeddings, enabling recovery of incomplete reasoning paths.

Exploration is initialized from a set of topic entities identified in the question, which serve as the starting points for all reasoning paths. We denote the initial topic entity set as $\mathcal{E}^{0} = \left\{ e^{0}_{1}, e^{0}_{2}, \ldots, e^{0}_{N_{0}} \right\}$, where $N_{0}$ is the number of topic entities in the question. Starting from $\mathcal{E}^{0}$, \sys incrementally expands reasoning paths relevant to the question. A reasoning path $p_n \in \mathcal{P}$ at iteration $D$ is defined as $p_n = \left\{ \left( e^{d}_{s,n}, r^{d}_{n}, e^{d}_{o,n} \right) \right\}_{d=1}^{D_{p_n}} .$ Let $\mathcal{E}^{D-1}$ denote the set of entities reached at iteration $D-1$: $\mathcal{E}^{D-1} = \left\{ e^{D-1}_{1}, e^{D-1}_{2}, \ldots, e^{D-1}_{N_{D-1}} \right\}.$
This set defines the starting point for the next exploration step, from which new candidate triples are retrieved or inferred.\looseness-2

The exploration phase consists of two complementary components: \textbf{relation exploration} and \textbf{entity exploration}. In each component, \sys retrieves candidate relations or entities, which are then filtered by an LLM based on their relevance to the question and the current reasoning context.

\subsubsection{Relation Exploration} Starting from a topic entity, exploration entails choosing a fitting relation. Instead of considering the full relation vocabulary, which would be prohibitively large, we restrict to a smaller set of relevant relations. Our approach relies on the observation that entity types exhibit characteristic patterns of associated relations~\cite{recoin}. To capture these patterns, we aggregate relation statistics at the entity-type level. For example, relations commonly associated with humans, such as \texttt{educated\_at} or \texttt{spouse}, differ from those associated with universities, such as \texttt{has\_chancellor}. Entity type information is obtained from the KG ontology when available (e.g., \texttt{P31:instance\_of} in Wikidata). For Freebase, we infer coarse entity types from relation namespace prefixes, e.g., entities participating in \texttt{film.film.directed\_by} are associated with the type \texttt{film.film}. Relation frequencies are then aggregated across entities sharing the same type, enabling sparse entities to benefit from type-level co-occurrence patterns.

\softbsubsec{Relation Selection} \sys first retrieves relations directly connected to $e_c$ in the KG. To mitigate graph incompleteness, this set is augmented with curated relations derived from corpus-level statistics. Specifically, we estimate the relevance of a relation $r_j \in \mathcal{R}$ to $e_c$ using a \textbf{type-level entity--relation co-occurrence matrix} $F$, where $F[t,j]$ denotes the frequency with which entities of type $t$ co-occur with relation $r_j$ in the KG. Relations are ranked using the score $\psi(r_j \mid e_c)$, where $t_c$ denotes the type of the current entity $e_c$:

\begin{equation}
\psi(r_j \mid e_c) = \frac{F[t_c, j]}{\sum_{k \in \mathcal{R}} F[t_c, k]},
\label{eq:relation_scoring}
\end{equation}

\begin{figure}[hb]
    \centering
    \includegraphics[width=\columnwidth]{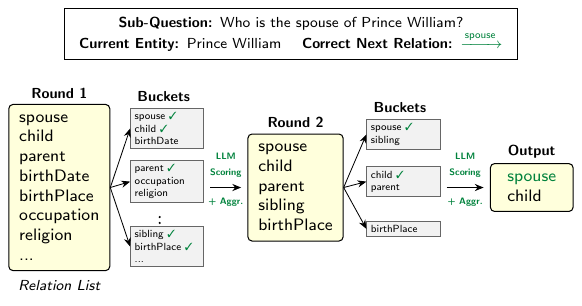}
    \caption{Hierarchical bucketing prompt strategy for relation exploration.}
    \label{fig:hierarchical_bucketing}
\end{figure}

\softbsubsec{LLM Relation Pruning}\label{sec:relation-pruning}
We form a candidate relation set $\mathcal{R}_{\text{cand}}(e_c)$ by combining relations observed in the KG with curated relations: $\mathcal{R}_{\text{cand}} = \mathcal{R}_{\text{KG}} \cup \mathcal{R}_{\text{cur}}$, where $\mathcal{R}_{\text{KG}}$ corresponds to the entity's neighboring relations and $\mathcal{R}_{\text{cur}}$ to its type-level co-occurrence relations, so that candidates remain structurally and semantically connected through the current entity and its associated entity type. If $\mathcal{R}_{\text{cand}}(e_c)$ exceeds a threshold (bucket size), using a single LLM prompt for pruning can become ineffective; hence, we introduce a novel \textbf{hierarchical bucketing prompt strategy} that partitions $\mathcal{R}_{\text{cand}}(e_c)$ into smaller buckets, prunes each independently with the LLM, and merges the retained relations across stages for a final refinement step that yields $\mathcal{R}(e_c)$. Fig.~\ref{fig:hierarchical_bucketing} illustrates this approach.

Relations are assigned to buckets at random; this is sufficient because the candidate pool is already restricted to structurally and semantically connected relations, and multi-retention with cross-stage merging recovers useful relations even when their bucketmates are unrelated. In each round, each bucket is independently evaluated by the LLM conditioned on the question $Q$, sub-questions $\{SQ_i\}$, and current topic entity $e_c$, which together provide contextual and structural information about the current reasoning step. Multiple highly relevant relations may be retained per bucket. This multi-retention is critical for bridge relations whose value only becomes apparent under longer compositions, avoiding premature commitment to a single locally dominant choice. The refined set $\mathcal{R}(e_c)$ is then used for subsequent entity exploration.

\subsubsection{Entity Exploration} Given the refined relation set from relation exploration, entity exploration identifies plausible next-hop entities that extend the current reasoning paths. At each step, for the current entity $e_c$ and a selected relation $r \in \mathcal{R}(e_c)$, candidate entities are explored using link prediction.

\softbsubsec{Link Prediction} For each $(e_c, r)$ pair, candidate entities are generated by scoring entities in the KG embedding space using the KGE scoring function $\phi$. Depending on relation direction, entities are ranked according to either $\phi(e_c, r, e)$ or $\phi(e, r, e_c)$. The top-$K$ ranked entities (\textbf{Entity Context K}) form the candidate set $\mathcal{E}_{\text{cand}}(e_c, r)$. This step allows \sys to recover plausible entities not explicitly connected in the incomplete KG.
When $\mathcal{E}_{\text{cand}}(e_c, r)$ becomes large, we use a lightweight pre-trained DistilBERT model~\cite{sanh2019distilbert} to score semantic similarity between the question $Q$ and candidate entities, filtering out irrelevant ones.

\softbsubsec{LLM Entity Pruning} Since embedding-based inference may introduce noise, \sys employs an LLM to prune candidate entities. The LLM is provided with $Q$, $e_c$, $r$, and $\mathcal{E}_{\text{cand}}(e_c, r)$, and outputs a refined entity set $\mathcal{E}(e_c, r)$. Each retained entity $e \in \mathcal{E}(e_c, r)$ extends the current reasoning path, yielding new candidate paths for subsequent exploration.
\subsection{Reasoning Phase}

Once candidate reasoning paths have been constructed, \sys enters the reasoning phase to decide whether to continue exploration or produce an answer. In this phase, the LLM reasons using the question, the current reasoning paths, and the maintained sub-question states.

\softbsubsec{Planning Status Update} Following PoG~\cite{chen2024plan}, the LLM is prompted with the question $Q$ and the currently explored reasoning paths $\mathcal{P}$ to update the status of each sub-question. Specifically, the LLM summarizes which sub-questions $SQ_i$ are answered by the current paths and updates their corresponding status, which guides subsequent exploration and reasoning iterations.

\softbsubsec{Evaluation} Based on the current reasoning paths $\mathcal{P}$ and sub-question status $\mathcal{S}$, the LLM evaluates whether sufficient evidence has been accumulated to infer an answer. If so, the LLM integrates the reasoning paths and sub-question status to identify the answer entity. Otherwise, if the maximum search depth has not been reached, the terminal entities of the current paths become new current entities $e_c$, and \sys returns to the exploration phase to gather additional evidence. This ensures exploration is guided by unresolved sub-questions and answers are grounded in the collected evidence.
\section{Experiments}


Our main evaluation examines \sys's performance, robustness, and cost under a standard KG incompleteness protocol. We then focus our analysis on component contributions and sensitivity to key design choices. We conclude by assessing generalization to a different KG and incompleteness setting.

\subsection{Experimental Setup}

\softbsubsec{Datasets}
We evaluate \sys on three multi-hop KGQA benchmarks: WebQSP~\cite{yih2016value}, CWQ~\cite{talmor2018web}, and BRINK~\cite{brink}. WebQSP and CWQ require multi-hop reasoning over the Freebase knowledge graph~\cite{bollacker2008freebase}, while BRINK evaluates KGQA over Wikidata5m~\cite{wang2021kepler} under both complete and incomplete graph settings.

\softbsubsec{Evaluation Metrics}
Following prior work \cite{chen2024plan,sun2023think},
we report exact match accuracy (Hits@1) on WebQSP and CWQ, where a prediction is correct if the predicted entity matches any gold answer entity. For BRINK, we report F1 following its original evaluation protocol ~\cite{brink}.

\softbsubsec{KG Incompleteness Protocol}
To evaluate robustness under incomplete KGs, for WebQSP and CWQ, we follow the protocol in GoG~\cite{xu2024generate} where the complete KG is denoted \textbf{CKG}, and four incomplete KG versions are created: \textbf{IKG-20\%}, \textbf{IKG-40\%}, \textbf{IKG-60\%}, and \textbf{IKG-80\%}, by randomly removing 20\%, 40\%, 60\%, and 80\% of the \emph{crucial triples} required to answer each question. Crucial triples lie along the gold reasoning path from the topic entity to the correct answer. All relations between the same entity pairs are also removed. For BRINK, we use its provided incomplete split. For each incomplete-KG setting, we train the KG embeddings on the corresponding pruned graph, excluding all removed triples.

\softbsubsec{Baselines}
We compare \sys against representative KGQA baselines spanning supervised and no task-specific KGQA training settings. As an LLM-only reference, we evaluate Chain-of-Thought (CoT)~\cite{wei2022chain} without access to the knowledge graph for each LLM backbone. \textit{LLM-driven semantic parsing methods} translate questions into executable KG queries; we compare against KB-BINDER~\cite{li2023few} and ChatKBQA~\cite{luo2023chatkbqa}. \textit{Supervised KGQA methods} use task-specific supervision for KG-based question answering; we include EmbedKGQA~\cite{saxena2020improving}. \textit{LLM-based KG reasoning methods} integrate LLMs with structured retrieval and graph reasoning; we compare against StructGPT~\cite{jiang2023structgpt}, RoG~\cite{luo2023reasoning}, ToG~\cite{sun2023think}, PoG~\cite{chen2024plan}, and GoG~\cite{xu2024generate}.

\softbsubsec{Training Regime} We use the term \emph{No Task-Specific KGQA Training} to describe the family of methods that do not train or fine-tune on KGQA supervision, including the use of question-answer pairs or annotated reasoning chains. Nonetheless, this designation permits the use of pretrained components, such as KG embeddings and small language models, as well as other auxiliary tools.

\softbsubsec{Default LLM}
Following prior work, we use GPT-3.5-Turbo as the default model. Alternatives are evaluated in Section~\ref{sec:overall} and Appendix~\ref{app:llms}.

\softbsubsec{Reproducibility}
Details on code, datasets, prompts, relevant parameters, and experimental setup are available on the project repository:\\
\url{https://github.com/colab-nyuad/XoG}
\subsection{Overall Performance Comparison}
\label{sec:overall}
Table~\ref{tab:kg_setting_results} reports results for WebQSP and CWQ under complete (CKG) and incomplete (IKG-40\%) settings. On complete graphs, \sys remains competitive with or outperforms comparable methods that do not use task-specific KGQA training. Fine-tuned models achieve strong performance but require task-specific training data. The performance gap between \sys and ToG highlights the limitations of traversal-based reasoning. In contrast, the relatively close performance of \sys and PoG suggests that \sys's information recovery provides comparatively limited additional benefit when the graph is complete, though, as shown below, this benefit grows substantially once the graph becomes incomplete.

\begin{table}[t!]
\centering

\resizebox{\columnwidth}{!}{%
\begin{threeparttable}
\begin{tabular}{l*{4}{>{\centering\arraybackslash}p{2cm}}}
\toprule
\textbf{Method}
& \multicolumn{2}{c}{\textbf{WebQSP}}
& \multicolumn{2}{c}{\textbf{CWQ}} \\
\cmidrule(lr){2-3} \cmidrule(lr){4-5}
& \textbf{CKG} & \textbf{IKG-40\%} & \textbf{CKG} & \textbf{IKG-40\%} \\
\midrule
\multicolumn{5}{c}{\textit{CoT (Without Knowledge Graph)}} \\
\midrule
GPT-3.5-Turbo& \multicolumn{2}{c}{76.0} & \multicolumn{2}{c}{54.1} \\
GPT-4     & \multicolumn{2}{c}{76.4} & \multicolumn{2}{c}{55.5} \\
GPT-5.5   & \multicolumn{2}{c}{79.1} & \multicolumn{2}{c}{74.0} \\
Claude Opus 4.8   & \multicolumn{2}{c}{80.7} & \multicolumn{2}{c}{62.8} \\
\midrule
\multicolumn{5}{c}{\textit{Supervised KGQA}} \\
\midrule

EmbedKGQA & 66.6 & 42.5 & 44.7 & NA \\
RoG\tnote{*}      & 88.6 & 78.2 & 66.1 & 54.2 \\
ChatKBQA\tnote{*} & 78.1 & 49.5 & 76.5 & 39.3 \\
\midrule
\multicolumn{5}{c}{\textit{No Task-Specific KGQA Training (GPT-3.5-Turbo)}} \\
\midrule
KB-BINDER\tnote{*} & 50.7 & 38.4 & -- & -- \\
StructGPT\tnote{*} & 76.4 & 60.1 & -- & -- \\
ToG\tnote{*}       & 76.9 & 63.4 & 47.2 & 37.9 \\
GoG\tnote{*}       & 78.7 & 66.6 & 55.7 & 44.3 \\
PoG\tnote{±}       & \underline{82.0} & \underline{69.7} & \underline{63.2} & \underline{53.7} \\
XoG          & \textbf{83.7} & \textbf{77.6} & \textbf{64.4} & \textbf{55.7} \\
\midrule
\multicolumn{5}{c}{\textit{No Task-Specific KGQA Training (GPT-4)}} \\
\midrule
ToG\tnote{*}       & 80.3 & 71.8 & 71.0 & 56.1 \\
GoG\tnote{*}       & 84.4 & \underline{80.3} & \textbf{75.2} & \underline{60.4} \\
PoG\tnote{±}       & \underline{85.2} & 71.6 & 71.9 & 58.0 \\
XoG          & \textbf{85.7} & \textbf{83.5} & \underline{72.0} & 64.7\\
\midrule
\multicolumn{5}{c}{\textit{No Task-Specific KGQA Training (GPT-5.5)}}\\
\midrule
ToG &\underline{86.2}	& 78.6	& \underline{78.0}	&\underline{74.1} \\
XoG & \textbf{87.3}	& \textbf{83.8}	& 80.5\rlap{\kern0.3em$^{\blacktriangle}$}	&75.9\rlap{\kern0.3em$^{\blacktriangle}$} \\
\midrule
\multicolumn{5}{c}{\textit{No Task-Specific KGQA Training (Claude Opus 4.8)}}\\
\midrule
ToG &\underline{85.8}	&78.6	&\underline{72.6}	&\underline{68.2} \\
XoG &89.2\rlap{\kern0.3em$^{\blacktriangle}$}	&85.4\rlap{\kern0.3em$^{\blacktriangle}$}	&\textbf{77.6}	&\textbf{68.8} \\
\bottomrule
\end{tabular}%

\begin{tablenotes}[flushleft]
\scriptsize
\item[*] Results are taken from GoG \cite{xu2024generate}.
\item[±] Results obtained using PoG \cite{chen2024plan} code on the incompleteness datasets.
\end{tablenotes}
\end{threeparttable}
}
\caption{Hits@1 scores on WebQSP and CWQ under complete (CKG) and incomplete (IKG-40\%) knowledge graph settings. CoT (No KG) is evaluated without KG access and serves as a backbone-matched LLM-only baseline. \textbf{Bold} indicates the best result and \underline{underlined} the second best within each \emph{``No Task-Specific KGQA Training''} paradigm. $\blacktriangle$ marks the best overall result in each column.}
\label{tab:kg_setting_results}
\end{table}
Under incomplete graphs, all KG-based methods experience performance degradation due to missing triples, though the size of this degradation varies by method and dataset. Traversal-based approaches such as ToG are particularly vulnerable to this setting, as broken reasoning paths can terminate exploration prematurely; PoG is designed to partially mitigate this through adaptive exploration and self-correction, which enables the discovery of alternative reasoning paths when the initial path is blocked. Compared to GoG, which relies on LLM-generated facts to fill in missing information, \sys instead adopts a graph-grounded recovery strategy, which we find is associated with consistently stronger performance under KG incompleteness.

We further evaluate \sys across multiple LLM backbones. For each backbone, we evaluate \emph{CoT} as a backbone-matched LLM-only baseline, while ToG represents iterative LLM-guided KG exploration. Across all evaluated backbones, \sys consistently outperforms both \emph{CoT} and ToG on WebQSP and CWQ under CKG, and continues to outperform ToG under IKG-40\% as backbones grow stronger, indicating that improved LLM reasoning alone does not eliminate the challenges introduced by missing graph information.

To assess whether KGQA performance is driven by LLM parametric memorization, we examine the \emph{CoT} setting, where the model must rely solely on its parametric knowledge. CoT accuracy alone cannot confirm or rule out memorization. However, \emph{CoT} falls well below \sys under both CKG and IKG-40\%, so parametric knowledge alone cannot reproduce the performance achieved with KG access, and \sys's gains are not primarily due to memorized Freebase facts during pretraining. This trend holds across additional LLM backbones (Appendix~\ref{app:llms}).

\subsection{Robustness to KG Sparsity}
\begin{figure}[!ht]
\centering

\resizebox{0.85\columnwidth}{!}{%
\begin{tikzpicture}
  \node[anchor=center, draw=black, inner sep=4pt, rounded corners=2pt] at (0,0) {
    \begin{tabular}{ccccc}
      \tikz{\draw[black!50, dotted, thick] (0,0) -- (0.4,0); \draw[black!50, mark=square, thick, mark options={solid, fill=black!50}] plot coordinates {(0.2,0)};} StructGPT &
      \tikz{\draw[blue!80!black, solid, thick] (0,0) -- (0.4,0); \draw[blue!80!black, mark=triangle*, thick] plot coordinates {(0.2,0)};} ToG &
      \tikz{\draw[violet, dashed, thick] (0,0) -- (0.4,0); \draw[violet, mark=diamond*, thick] plot coordinates {(0.2,0)};} GoG &
      \tikz{\draw[orange!80!black, dashdotted, thick] (0,0) -- (0.4,0); \draw[orange!80!black, mark=*, thick] plot coordinates {(0.2,0)};} PoG &
      \tikz{\draw[green!60!black, solid, thick] (0,0) -- (0.4,0); \draw[green!60!black, mark=star, mark size=3.5pt, thick] plot coordinates {(0.2,0)};} \textbf{XoG}
    \end{tabular}
  };
\end{tikzpicture}%
}

\vspace{0.3em}

\resizebox{\columnwidth}{!}{%
\begin{tikzpicture}

\begin{axis}[
  name=webqsp,
  width=5.8cm, height=4.5cm,
  xmin=-0.3, xmax=4.3,
  xtick={0,1,2,3,4},
  xticklabels={CKG, 20, 40, 60, 80},
  ymajorgrids,
  xlabel={Missing Triple Rate (\%)},
  ylabel={Hits@1 (\%)},
  title={\textbf{(a) WebQSP}},
  ymin=40, ymax=92,
  ytick={40,50,60,70,80,90},
  line width=1pt, mark size=2pt,
]
  \addplot[mark=square, draw=black!50, dotted, thick, mark options={solid, fill=black!50}] coordinates {
    (0,76.0) (1,67.8) (2,60.1) (3,51.7) (4,43.7)
  };

  \addplot[mark=triangle*, draw=blue!80!black, solid, thick, mark options={fill=blue!80!black}] coordinates {
    (0,76.9) (1,70.3) (2,61.4) (3,60.6) (4,55.9)
  };

  \addplot[mark=diamond*, draw=violet, dashed, thick, mark options={fill=violet}] coordinates {
    (0,78.7) (1,70.8) (2,66.6) (3,62.6) (4,56.5)
  };

  \addplot[mark=*, draw=orange!80!black, dashdotted, thick, mark options={fill=orange!80!black}] coordinates {
    (0,81.5) (1,71.9) (2,69.7) (3,60.8) (4,60.7)
  };

  \addplot[mark=star, draw=green!60!black, solid, thick, mark size=3.5pt, mark options={fill=green!60!black}] coordinates {
    (0,83.7) (1,80.4) (2,77.6) (3,73.8) (4,71.4)
  };
\end{axis}

\begin{axis}[
  at={(webqsp.east)},
  anchor=west,
  xshift=0.6cm,
  width=5.8cm, height=4.5cm,
  xmin=-0.3, xmax=4.3,
  xtick={0,1,2,3,4},
  xticklabels={CKG, 20, 40, 60, 80},
  ymajorgrids,
  xlabel={Missing Triple Rate (\%)},
  title={\textbf{(b) CWQ}},
  ymin=25, ymax=72,
  ytick={30,40,50,60,70},
  line width=1pt, mark size=2pt,
]
  \addplot[mark=triangle*, draw=blue!80!black, solid, thick, mark options={fill=blue!80!black}] coordinates {
    (0,47.2) (1,40.5) (2,37.9) (3,33.7) (4,31.4)
  };

  \addplot[mark=diamond*, draw=violet, dashed, thick, mark options={fill=violet}] coordinates {
    (0,55.7) (1,44.9) (2,44.3) (3,36.2) (4,34.4)
  };

  \addplot[mark=*, draw=orange!80!black, dashdotted, thick, mark options={fill=orange!80!black}] coordinates {
    (0,63.8) (1,57.8) (2,53.7) (3,50.5) (4,48.4)
  };

  \addplot[mark=star, draw=green!60!black, solid, thick, mark size=3.5pt, mark options={fill=green!60!black}] coordinates {
    (0,64.4) (1,59.1) (2,55.7) (3,52.9) (4,50.7)
  };
\end{axis}

\end{tikzpicture}%
}

\caption{Hits@1 results on complete (CKG) and incomplete (IKG-\%) knowledge graphs.}
\label{fig:sparsity}
\end{figure}
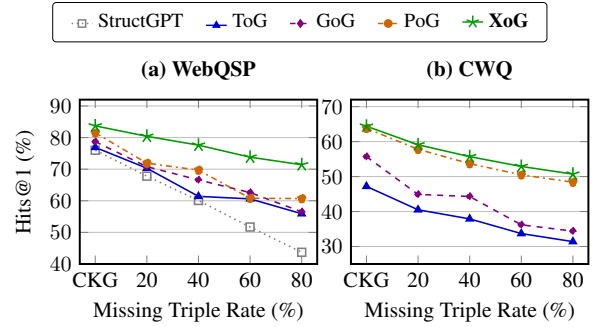

Next, we analyze robustness under progressively increasing KG sparsity. We focus on WebQSP and CWQ, whose controlled incompleteness protocol enables evaluation across multiple sparsity levels (IKG-20\%--80\%) as shown in Fig.~\ref{fig:sparsity}. We observe that planning-based methods such as PoG and \sys achieve similar performance when the KG is complete. As more relations are removed, the performance of all methods gradually degrades. Notably, traversal-based approaches such as ToG exhibit the sharpest accuracy drop because they depend on access to the full KG structure. PoG is more robust due to its reflection module, but suffers noticeable degradation under severe incompleteness. GoG partially compensates for missing links using LLM-generated knowledge, though its gains diminish under extreme sparsity, particularly on CWQ. In contrast, XoG consistently achieves the strongest performance across all sparsity levels on both datasets, suggesting that its augmented exploration strategy supports more robust reasoning as graph incompleteness increases.

\subsection{Efficiency Analysis}\label{sec:llm-usage}
\begin{table*}[ht!]
\centering
\setlength{\abovecaptionskip}{1pt}
\vspace{4pt}
\setlength{\tabcolsep}{2.5pt}
\renewcommand{\arraystretch}{1}
\small
\resizebox{\textwidth}{!}{%
\begin{tabular}{@{}llrrrrrrrrrrrr@{}}
\toprule
& & \multicolumn{3}{c}{\textbf{LLM Calls}} & \multicolumn{3}{c}{\textbf{Input Tokens}} & \multicolumn{3}{c}{\textbf{Output Tokens}} & \multicolumn{3}{c}{\textbf{Total Tokens}} \\
\cmidrule(lr){3-5} \cmidrule(lr){6-8} \cmidrule(lr){9-11} \cmidrule(lr){12-14}
\textbf{Dataset} & \textbf{Sparsity} & \textbf{PoG} & \textbf{XoG} & \textbf{$\Delta$} & \textbf{PoG} & \textbf{XoG} & \textbf{$\Delta$} & \textbf{PoG} & \textbf{XoG} & \textbf{$\Delta$} & \textbf{PoG} & \textbf{XoG} & \textbf{$\Delta$} \\
\midrule
\multirow{3}{*}{CWQ}
    & CKG      & 14.67 & \textbf{13.59} & \deltL{$-$7\%}  & 8,852  & \textbf{8,192}  & \deltL{$-$7\%}  & 390 & \textbf{361} & \deltL{$-$7\%}  & 9,243  & \textbf{8,553}  & \deltL{$-$7\%} \\
    & IKG-40\% & 17.68 & \textbf{14.00} & \deltM{$-$21\%} & 10,726 & \textbf{8,397}  & \deltM{$-$22\%} & 464 & \textbf{373} & \deltM{$-$20\%} & 11,191 & \textbf{8,770}  & \deltM{$-$22\%} \\
    & IKG-80\% & 19.50 & \textbf{14.30} & \deltH{$-$27\%} & 11,886 & \textbf{8,638}  & \deltH{$-$27\%} & 506 & \textbf{386} & \deltM{$-$24\%} & 12,393 & \textbf{9,024}  & \deltH{$-$27\%} \\
\midrule
\multirow{3}{*}{WebQSP}
    & CKG      & 9.11  & \textbf{8.90}  & \deltL{$-$2\%}  & 5,391  & \textbf{5,160}  & \deltL{$-$4\%}  & 286 & \textbf{271} & \deltL{$-$5\%}  & 5,677  & \textbf{5,432}  & \deltL{$-$4\%} \\
    & IKG-40\% & 13.70 & \textbf{9.20}  & \deltH{$-$33\%} & 8,033  & \textbf{5,376}  & \deltH{$-$33\%} & 376 & \textbf{258} & \deltH{$-$31\%} & 8,409  & \textbf{5,634}  & \deltH{$-$33\%} \\
    & IKG-80\% & 13.22 & \textbf{9.40}  & \deltH{$-$29\%} & 7,812  & \textbf{5,447}  & \deltH{$-$30\%} & 368 & \textbf{255} & \deltH{$-$31\%} & 8,180  & \textbf{5,703}  & \deltH{$-$30\%} \\
\bottomrule
\end{tabular}%
}
\caption{LLM usage comparison between PoG and XoG across datasets and KG sparsity versions.}
\label{tab:efficiency}
\end{table*}
We study the efficiency of \sys by comparing it with PoG, the most closely related baseline in pipeline structure and performance. We measure the average number of LLM calls and the token consumption required to answer a question (Table~\ref{tab:efficiency}).
Across all settings, \sys consistently requires fewer LLM calls and substantially fewer tokens than PoG. On complete graphs, both methods exhibit similar LLM usage, with \sys showing modest reductions (2--7\%). However, as sparsity increases, PoG's iterative reflection mechanism leads to increasing LLM calls and token consumption, while \sys remains stable. Under incomplete settings, \sys reduces total token consumption by up to 33\% compared to PoG.
This efficiency gain stems from \sys's integration of selection tools such as relation selection and embedding-based entity retrieval. By recovering relevant evidence early, \sys avoids repeated exploration and reflection cycles and reserves LLM usage primarily for pruning and final reasoning. As a result, XoG achieves a superior accuracy–cost trade-off and scales more gracefully as KG incompleteness increases.
End-to-end inference takes approximately 12 seconds per question, with LLM inference accounting for roughly 85\% of this time; KGE scoring adds only about 0.3 seconds, with the remainder spent on SPARQL querying, relation-expansion lookups, and DistilBERT filtering. This indicates that LLM usage remains the dominant inference cost, while the added embedding-based recovery introduces relatively little overhead.

\subsection{Sensitivity Analysis}
\label{app:sensitivity}
Next, we analyze the sensitivity of \sys to key design choices: hierarchical relation selection and the quality of embedding-based entity retrieval. We use WebQSP for these analyses because its linear reasoning paths provide a clear sequence of relation and entity selections, making the behavior of the individual modules easier to interpret.

\softbsubsec{The Effect of Hierarchical Relation Selection and Context Size} To evaluate the hierarchical relation selection strategy in isolation, we measure first-hop relation prediction accuracy under varying KG sparsity levels and bucket sizes. Table~\ref{tab:bucket_size} shows that hierarchical relation selection improves performance under both complete and incomplete KG settings. A bucket size of 30 achieves the best results across all sparsity levels while maintaining stable performance under increasing KG sparsity. The candidate relation set contains approximately 98 relations per query, including both neighboring KG relations and additional relations introduced through type-level co-occurrence statistics. As bucket size increases, performance gradually declines and drops further when hierarchy is removed entirely. These results suggest that hierarchical bucketing benefits relation exploration by enabling progressive refinement over manageable subsets of the candidate space rather than requiring the LLM to reason over the full relation set simultaneously. 
\begin{table}[h!]
\centering
\small
\setlength{\tabcolsep}{5pt}
\begin{tabular}{lccccc}
\toprule
& \multicolumn{5}{c}{\textbf{KG Sparsity (\%)}} \\
\cmidrule(lr){2-6}
\textbf{Bucket Size} & \textbf{CKG} & \textbf{20} & \textbf{40} & \textbf{60} & \textbf{80} \\
\midrule
30           & \textbf{80.40} & \textbf{80.62} & \textbf{80.60} & \textbf{80.12} & \textbf{80.22} \\
40           & 80.32 & 80.12 & 80.32 & 80.02 & 79.82 \\
50           & 79.31 & 78.60 & 79.10 & 78.50 & 78.80 \\
70           & 77.99 & 78.29 & 77.89 & 78.19 & 76.98 \\
No Hierarchy & 77.38 & 77.28 & 77.08 & 77.28 & 76.97 \\
\bottomrule
\end{tabular}
\caption{Effect of bucket size on relation selection hit rate on WebQSP under different KG sparsity levels. Columns are CKG (complete knowledge graph), and missing-triples rates (IKG-\%).}
\label{tab:bucket_size}
\end{table}

\softbsubsec{The Effect of Varying Retrieval Quality} To characterize XoG's dependence on link-prediction quality, we conduct a synthetic study of how retrieval quality affects downstream entity selection. We simulate retrieval quality levels by varying the Hits@100 of the candidate entity set during first-hop entity selection on WebQSP and evaluate the LLM-based selection module under each setting.

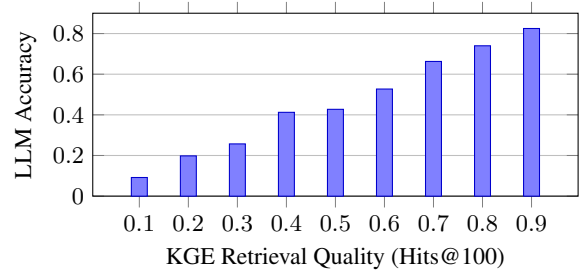
\begin{figure}[h!]
\centering
\footnotesize
\begin{tikzpicture}
\begin{axis}[
  width=8cm, height=4cm,
  ybar,
  bar width=6pt,
  xmin=0.05, xmax=0.95,
  xtick={0.1,0.2,0.3,0.4,0.5,0.6,0.7,0.8,0.9},
  xticklabel style={/pgf/number format/.cd,fixed,precision=1},
  xlabel={KGE Retrieval Quality (Hits@100)},
  ylabel={LLM Accuracy},
  ymin=0, ymax=0.9,
  ytick={0,0.2,0.4,0.6,0.8},
  ymajorgrids,
  enlarge x limits=0.05,
]
  \addplot[draw=blue!80!black, fill=blue!50!white] coordinates {
    (0.10, 0.0917)
    (0.20, 0.198)
    (0.30, 0.257)
    (0.40, 0.4124)
    (0.50, 0.4272)
    (0.60, 0.527)
    (0.70, 0.6628)
    (0.80, 0.7398)
    (0.90, 0.8249)
  };
\end{axis}
\end{tikzpicture}%
\caption{Effect of retrieval quality (Hits@100) on LLM entity selection accuracy.}
\label{fig:hits100_llm_accuracy}
\end{figure}

Figure~\ref{fig:hits100_llm_accuracy} shows that LLM entity selection accuracy closely follows the retrieval quality. As Hits@100 decreases, the LLM accuracy also declines, suggesting that the LLM is generally effective at identifying the correct entity when it is present within the retrieved candidate pool. Thus, entity exploration is primarily constrained by embedding-based retrieval quality rather than LLM selection. This is consistent with the error analysis in Appendix~\ref{app:error}, where retrieval gaps dominate XoG's errors and increase with KG sparsity.

\subsection{Ablation Study}\label{sec:ablation}
We complement the module-level analysis with end-to-end ablations to assess component contributions to KGQA performance.
Table~\ref{tab:ablation_xog} reports Hits@1 on WebQSP under complete and incomplete KG settings.
The ablations test two axes: recovery mechanisms (co-occurrence statistics and link prediction) and LLM-guided selection (relation and entity pruning). A detailed standalone evaluation of the relation and entity selection modules is provided in Appendix~\ref{sec:ablation}.

\begin{table}[h]
\centering

\resizebox{\linewidth}{!}{
\begin{tabular}{lccc}
\toprule
Method & CKG & IKG-40\% & IKG-80\% \\
\midrule
XoG (full) & \textbf{83.7} & \textbf{77.6} & \textbf{72.3} \\
\midrule
\multicolumn{4}{l}{\textit{(a) Recovery mechanisms}} \\
\quad w/o Relation Selection & 82.9 & 73.8 & 68.9 \\
\quad w/o Link Prediction     & 76.2 & 62.0 & 67.0 \\
\midrule
\multicolumn{4}{l}{\textit{(b) LLM-guided selection}} \\
\quad w/o LLM Entity Prune (Embed top1)     & 77.9 & 69.8 & 65.7 \\
\quad w/o LLM Entity Prune (Embed top3)     &  78.6& 73.2 &66.9  \\
\quad w/o LLM Relation Prune (DistilBERT)     &  78.2& 74.6 &70.7  \\

\bottomrule
\end{tabular}
}
\caption{Ablation results of XoG under different sparsity levels on WebQSP.}
\label{tab:ablation_xog}
\end{table}

\softbsubsec{Recovery mechanisms}
The \textit{w/o Relation Selection} version removes the curated relation dictionary and relies only on adjacent relations, while \textit{w/o Link Prediction} disables embedding-based link prediction, restricting entity discovery to existing graph neighbors. Removing relation selection has minimal effect when the graph is complete, but causes a noticeable decline on sparser graphs, highlighting its importance when neighborhood information is incomplete. Removing link prediction yields a larger drop across all settings, including CKG. We hypothesize that the benefit under CKG occurs because link prediction can also help disambiguate among multiple candidate entities for many-to-many relations (e.g., siblings and co-actors), where graph neighbors alone may be insufficient for fine-grained selection. The non-monotonicity between IKG-40\% and IKG-80\% may arise because performance depends not only on the amount of missing information, but also on which triples are removed and which alternative reasoning paths remain available.

\softbsubsec{LLM-guided selection}
During exploration, \sys uses the LLM to prune candidate relations and entities based on the current reasoning context. To assess whether its use improves upon simpler selection strategies, Table~\ref{tab:ablation_xog} compares the full model with two types of ablation: \emph{w/o LLM Entity Prune}, where we retain the top-1 or top-3 embedding-scored entity candidates, and \emph{w/o LLM Relation Prune}, where we select relations using a DistilBERT-based similarity baseline. All three selection alternatives consistently degrade performance, indicating that the LLM's selections improve \sys accuracy over these alternatives.

\subsection{Generalization to Wikidata-based KGQA}

Finally, to assess whether \sys's effectiveness extends beyond Freebase-based benchmarks, we evaluate it on BRINK: a Wikidata5m-based benchmark. This provides a complementary test under a different incompleteness protocol, which removes rule-mined triples while ensuring that the questions can still be answered. Table~\ref{tab:brink} shows that under incompleteness, \sys achieves the highest F1 among methods that do not use task-specific KGQA training. Although supervised approaches such as RoG and GNN-RAG remain more robust, \sys substantially narrows this gap without task-specific supervision. On complete graphs, StructGPT performs best among the non-task-trained methods, whereas \sys performs similarly to PoG. This mirrors the Freebase results: \sys offers limited gains on complete graphs but consistently outperforms comparable baselines under KG incompleteness.

\begin{table}[t]
\centering

\resizebox{\columnwidth}{!}{%
\begin{tabular}{l@{\hskip 1.4cm}cc}
\toprule
\textbf{Method} & \textbf{Complete} & \textbf{Incomplete} \\
\midrule
\multicolumn{3}{c}{\textit{Supervised KGQA}} \\
\midrule
G-Retriever & 0.32 & 0.30 \\
RoG         & 0.78 & 0.62 \\
GNN-RAG     & 0.79 & 0.68\rlap{\kern0.3em$^{\blacktriangle}$} \\
\midrule
\multicolumn{3}{c}{\textit{No Task-Specific KGQA Training (GPT-3.5-Turbo)}} \\
\midrule
PoG         & \underline{0.71} & 0.34 \\
ToG         & 0.64 & 0.32 \\
StructGPT   & 0.81\rlap{\kern0.3em$^{\blacktriangle}$} & \underline{0.40} \\
XoG  & \underline{0.71} & \textbf{0.52} \\
\bottomrule
\end{tabular}%
}
\caption{F1 on the BRINK Wikidata benchmark using entity text labels, under complete and incomplete KG settings. Baseline numbers are taken from \citet{brink}. \textbf{Bold} indicates the best result and \underline{underlined} the second best within the \emph{``No Task-Specific KGQA Training''} paradigm. $\blacktriangle$ marks the best overall result in each column.}
\label{tab:brink}
\end{table}

\section{Conclusion}
We presented \sys, a framework for LLM-based question answering over incomplete knowledge graphs. By integrating graph-grounded recovery into an iterative \emph{planning--exploration--reasoning} paradigm, \sys recovers missing reasoning paths while grounding LLM reasoning in graph evidence. Extensive experiments demonstrate that this design improves both answer accuracy and token efficiency, consistently outperforming methods without task-specific KGQA training under KG incompleteness while requiring fewer LLM calls than comparable approaches. These gains hold across six LLM backbones and extend beyond Freebase-based evaluation to Wikidata, showing that the benefits of graph-grounded recovery persist across different backbone strengths and KG settings. The advantage grows with KG sparsity, and our error analysis shows that the remaining failures are dominated by retrieval gaps rather than LLM selection errors, highlighting improved retrieval and KG completion as key directions for further gains.
\section*{Limitations}\label{sec:limitations}

Our framework relies on a knowledge graph embedding (KGE) module for link prediction. In this work, we employ ComplEx as a strong, widely implemented, and computationally efficient baseline, which allows us to focus on analyzing the behavior of the proposed method. Nonetheless, our approach is model-agnostic by design and can readily incorporate more expressive or higher-accuracy knowledge graph completion models. To characterize the dependence of XoG on the underlying KGE component, we conduct a synthetic link-prediction sensitivity study in Section~\ref{app:sensitivity}.

We evaluate under synthetic incompleteness using the benchmark introduced by GoG~\cite{xu2024generate}, which removes crucial triples at incremental rates. 
This setup enables a fair comparison with existing baselines without task-specific KGQA training, but may not fully capture real-world incompleteness patterns, such as uneven coverage across entity classes~\cite{luggen2019nonparametric} or gaps in coverage of long-tail entities~\cite{tonon2016voldemortkg}.
We additionally validate on BRINK, a Wikidata-based benchmark; however, its incomplete split is also synthetically constructed through rule-mined triple removal. Establishing broader incomplete-KGQA benchmarks that better reflect naturally occurring incompleteness and account for KG evolution~\cite{difallah2025wikirag} remains an important direction for future work.

Our primary evaluation uses WebQSP and CWQ, which are grounded in Freebase and enable direct comparison with prior incomplete-KGQA baselines. We additionally evaluate on BRINK over Wikidata5m, providing evidence that XoG generalizes beyond Freebase. Nevertheless, broader generalization to other KGs and domains remains to be established. Applying XoG to a new KG requires access to entity-type information for constructing type-level relation co-occurrence statistics. Performance on KGs without accessible type structure therefore remains a limitation.

\bibliography{custom}
\appendix

\section{Overall Efficiency Analysis}\label{app:efficiency}

\sys relies on two offline preprocessing components that are computed once per KG setting, and therefore can be amortized over all inference instances. First, we train ComplEx KG embeddings using LibKGE~\cite{DBLP:conf/emnlp/BroscheitRKBG20}; on Freebase, this step takes approximately 5 hours on a single NVIDIA A100 GPU. Second, \sys constructs the type-level entity--relation co-occurrence matrix $F$ used for relation expansion in Eq.~\ref{eq:relation_scoring}. This requires a single pass over the training triples to accumulate counts between entity types and relations, followed by normalization over relations for each entity type; both steps are linear in their inputs. At inference time, \sys combines LLM calls with lightweight non-LLM components, with overall efficiency primarily determined by LLM usage, prompt length, and the number of exploration iterations.

\section{Error Analysis}\label{app:error}

\subsection{Retrieval and Reasoning Error Decomposition}
We split \sys's wrong predictions into two categories to attribute failures to the corresponding component:
\begin{itemize}[noitemsep,topsep=2pt,parsep=0pt,leftmargin=1.2em]
    \item \textbf{Selection Error}: the share of wrong predictions where the correct answer was present in the retrieved chains, but the LLM failed to select it. This isolates reasoner-side failures.
    \item \textbf{Retrieval Gap}: the share of wrong predictions where the correct answer was absent from the retrieved chains altogether. This isolates retriever-side failures caused by missing or unretrievable graph evidence.
\end{itemize}
As shown in Table~\ref{tab:error_analysis}, retrieval gaps dominate the error distribution across both datasets and grow further as sparsity increases, while selection errors remain comparatively low. This indicates that performance degradation under KG incompleteness is driven primarily by the quality of the embedding space used for retrieval, rather than by the LLM misusing the chains it does retrieve.
\begin{table}[h!]
\centering
\small
\setlength{\tabcolsep}{5pt}
\resizebox{\linewidth}{!}{
\begin{tabular}{llccc}
\toprule
\textbf{Dataset} & \textbf{Metric} & \textbf{CKG} & \textbf{IKG-40\%} & \textbf{IKG-80\%} \\
\midrule

\multirow{3}{*}{WebQSP}
& Accuracy& 83.7 & 77.6 & 72.3 \\
& Selection Error & 15.9 & 10.3 & 6.6 \\
& Retrieval Gap & 84.1 & 89.7 & 93.4 \\
\midrule

\multirow{3}{*}{CWQ}
& Accuracy& 64.4 & 55.7 & 48.7 \\
& Selection Error& 13.9 & 10.7 & 6.5 \\
& Retrieval Gap& 86.1 & 89.3 & 93.5 \\
\bottomrule
\end{tabular}
}
\caption{Error Analysis of XoG under increasing KG incompleteness (all values in \%).}
\label{tab:error_analysis}
\end{table}

\subsection{Robustness and Efficiency vs.\ PoG}

We next compare \sys against PoG on two axes: how often each system's answer is backed by a retrieved chain (i.e., a \emph{graph-grounded answer}), rather than produced from the LLM memory; and the \emph{average chain length}, which reflects how much exploration each system needs.

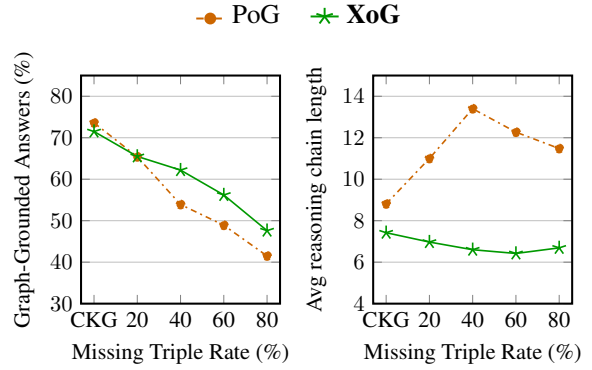
\begin{figure}[t]
\centering

\resizebox{0.42\columnwidth}{!}{%
\begin{tikzpicture}
  \node[anchor=center, inner sep=2pt] at (0,0) {
    \begin{tabular}{cc}
      \tikz{\draw[orange!80!black, dashdotted, thick] (0,0) -- (0.4,0); \draw[orange!80!black, mark=*, thick] plot coordinates {(0.2,0)};} PoG &
      \tikz{\draw[green!60!black, solid, thick] (0,0) -- (0.4,0); \draw[green!60!black, mark=star, mark size=3.5pt, thick] plot coordinates {(0.2,0)};} \textbf{XoG}
    \end{tabular}
  };
\end{tikzpicture}%
}

\vspace{0.1em}

\resizebox{\columnwidth}{!}{%
\begin{tikzpicture}

\begin{axis}[
  name=grounded,
  width=5.0cm, height=5.5cm,
  xmin=-0.3, xmax=4.3,
  xtick={0,1,2,3,4},
  xticklabels={CKG, 20, 40, 60, 80},
  ymajorgrids,
  xlabel={Missing Triple Rate (\%)},
  ylabel={Graph-Grounded Answers (\%)},
  ymin=30, ymax=85,
  ytick={30,40,50,60,70,80},
  line width=1pt, mark size=2pt,
]
  \addplot[mark=*, draw=orange!80!black, dashdotted, thick, mark options={fill=orange!80!black}] coordinates {
    (0,73.6) (1,65.4) (2,53.9) (3,48.9) (4,41.5)
  };

  \addplot[mark=star, draw=green!60!black, solid, thick, mark size=3.5pt, mark options={fill=green!60!black}] coordinates {
    (0,71.5) (1,65.5) (2,62.2) (3,56.2) (4,47.6)
  };
\end{axis}

\begin{axis}[
  at={(grounded.east)},
  anchor=west,
  xshift=1.6cm,
  width=5.0cm, height=5.5cm,
  xmin=-0.3, xmax=4.3,
  xtick={0,1,2,3,4},
  xticklabels={CKG, 20, 40, 60, 80},
  ymajorgrids,
  xlabel={Missing Triple Rate (\%)},
  ylabel={Avg reasoning chain length},
  ymin=4, ymax=15,
  ytick={4,6,8,10,12,14},
  line width=1pt, mark size=2pt,
]
  \addplot[mark=*, draw=orange!80!black, dashdotted, thick, mark options={fill=orange!80!black}] coordinates {
    (0,8.81) (1,11.00) (2,13.40) (3,12.27) (4,11.48)
  };

  \addplot[mark=star, draw=green!60!black, solid, thick, mark size=3.5pt, mark options={fill=green!60!black}] coordinates {
    (0,7.42) (1,6.97) (2,6.60) (3,6.42) (4,6.69)
  };
\end{axis}

\end{tikzpicture}%
}

\caption{PoG vs.\ XoG under varying KG sparsity on WebQSP. \textbf{(a)} Graph-grounded answer rate. \textbf{(b)} Average reasoning chain length.}
\label{fig:pog_vs_xog}
\end{figure}

Fig.~\ref{fig:pog_vs_xog}(a) reports the portion of \emph{graph-grounded answers} as the KG gets sparsified. For CKG the two methods behave similarly. As more triples are removed, PoG's dependence on direct traversal causes its grounded rate to drop, while \sys degrades more gracefully. This shows that embedding-based recovery lets \sys keep grounding its answers in the graph when paths are broken.

Fig.~\ref{fig:pog_vs_xog}(b) reports the average chain length among graph-grounded answers. \sys keeps its reasoning chains short across all sparsity levels, whereas PoG's chains become longer under incompleteness as it explores laterally to compensate for the missing links.

\section{Extended LLM Backbone Comparison}\label{app:llms}

Table~\ref{tab:change_llm} reports results for \sys with six backbone LLMs drawn from several model families and generations: GPT-3.5-Turbo, GPT-4, GPT-5.5, Qwen3-32B~\cite{yang2025qwen3}, LLaMA-3.3-70B~\cite{grattafiori2024llama}, and Claude Opus~4.8~\cite{anthropic2026opus48}. The choice of backbone has a large effect on performance under both complete and incomplete KGs, and no single model is best on both datasets. Claude Opus~4.8 is the strongest on WebQSP, while GPT-5.5 leads on CWQ. The gap between backbones is wider on CWQ. This may stem from the complex compositional nature of CWQ questions, which rely more heavily on question decomposition and sub-question planning; an area where stronger backbones can have an edge.

\begin{table}[h!]
\centering
\resizebox{\linewidth}{!}{
\begin{tabular}{llcccccc}
\toprule
& Setting & GPT-3.5 & Qwen3 & Llama-3.3 & GPT-4 & GPT-5.5 & Opus 4.8\\
\midrule
\multirow{3}{*}{\rotatebox[origin=c]{90}{WebQSP}}
  & CKG      & 83.7 & 77.9 & 84.9 & {85.7} & {87.3}& {89.2}\\
  & IKG-40\% & 77.6 & 71.8 & 81.6 & {83.5} &{83.8}&{85.4}\\
  & CoT      & 76.0 & 62.7 & {76.9} & 76.4 &{79.1}&{80.7}\\
\midrule
\multirow{3}{*}{\rotatebox[origin=c]{90}{CWQ}}
  & CKG      & 64.4 & 60.6 & 66.1 & {72.0}&{80.5}& {77.6}\\
  & IKG-40\% & 55.7 & 53.0 & 61.3 & {64.7}&{75.9} &{68.8}\\
  & CoT      & 54.1 & 48.0 & {57.5} & 55.5&{74.0}&{62.8} \\
\bottomrule
\end{tabular}
}
\caption{Hits@1 (\%) of XoG with different backbone LLMs across complete (CKG), incomplete (IKG-40\%), and CoT (No KG) settings.}
\label{tab:change_llm}
\end{table}

We also find that strong reasoning from parametric knowledge alone does not necessarily lead to strong reasoning over a KG. For example, in the CoT(No KG) setting, LLaMA-3.3 slightly outperforms GPT-4 on both WebQSP and CWQ. With KG access, however, GPT-4 comes out ahead on both. The benefit of KG access also varies across backbones, which ssuggests that some models use external knowledge more effectively than others. Finally, \sys remains effective across all six backbones under KG incompleteness. Although performance decreases from the complete KG to IKG-40\% for every model, the IKG-40\% results remain well above the corresponding CoT baselines. The robustness of \sys to missing facts is therefore not limited to a particular LLM family.

\section{Relation and Entity Selection Analysis}\label{sec:ablation}
We evaluate the relation and entity selection modules individually. Since both modules rely on LLM pruning, we report \emph{hit rate}, defined as the proportion of queries for which the correct relation or entity is present in the LLM-returned candidate set. All experiments are conducted on WebQSP versions, as the questions' linear reasoning paths enable controlled analysis of the selection modules.

\begin{table}[h!]
\centering
\begin{threeparttable}
\resizebox{\linewidth}{!}{%
\begin{tabular}{lccc}
\toprule
\textbf{Method} & \textbf{CKG} & \textbf{IKG-40\%} & \textbf{IKG-80\%} \\
\midrule
All Relations & 6.20 & 6.20 & 6.20 \\
Neighbors only & \textbf{80.20} & 35.00 & 10.54 \\
Relation Selection & 77.99 & \textbf{77.89} & \textbf{76.98} \\
\bottomrule
\end{tabular}%
}
\end{threeparttable}
\caption{Relation selection hit rate on WebQSP.}
\label{tab:webqsp_sparsity_predicates}
\end{table}

\softbsubsec{Relation Selection Module} To carry out this isolated evaluation, we apply relation selection to the question's topic entity to retrieve $K=70$ candidate relations, followed by hierarchical LLM pruning using the question and topic entity as context. Table~\ref{tab:webqsp_sparsity_predicates} shows that on the complete graph (CKG), the Neighbors baseline achieves the highest hit rate  by leveraging existing entity connections. However, as sparsity increases, this approach's performance degrades since relevant relations are proportionally missing from the neighborhood. In contrast, our relation selection module maintains stable hit rate across all sparsity levels, demonstrating effectiveness in recommending correct relations despite the graph incompleteness.

\begin{figure}[h]
  \centering
    \includegraphics[width=\columnwidth]{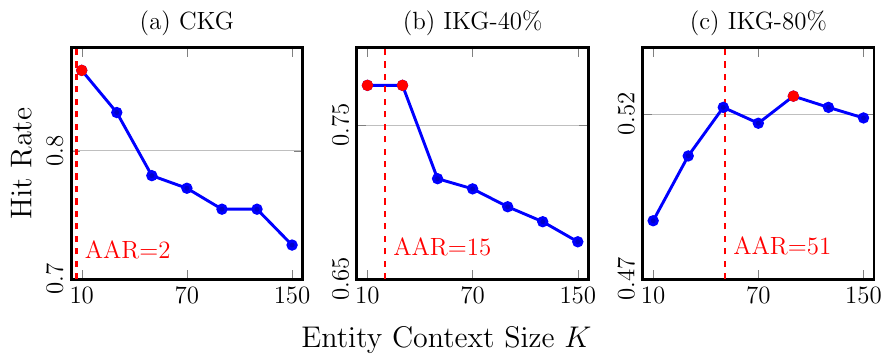}
  \caption{Effect of entity context size $K$ on entity selection hit rate. Average answer rank (AAR) is indicated.}
  \label{fig:entity-pruning-firsthop}

\end{figure}

\FloatBarrier
\softbsubsec{Entity Selection Module}
We apply link prediction to each question's (topic entity, ground-truth first-hop relation) pair to retrieve the top-$K$ candidate entities, followed by LLM pruning using the question as context. Fig.~\ref{fig:entity-pruning-firsthop} shows that on the complete graph (CKG), where the correct entity ranks near the top (average answer rank, AAR=2), a small context ($K=10$) achieves the best hit rate; increasing $K$ only introduces noise. As sparsity increases, link prediction quality degrades and average answer rank increases to 15 for IKG-40\% and 51 for IKG-80\%, requiring larger context sizes to ensure coverage of the correct entity. Ultimately, expanding $K$ introduces noise with larger context sizes.

\end{document}